\documentclass[letterpaper]{article} 
\usepackage{aaai23}  
\usepackage{times}  
\usepackage{helvet}  
\usepackage{courier}  
\usepackage[hyphens]{url}  
\usepackage{graphicx} 
\usepackage{natbib}  
\usepackage{caption} 
\usepackage{algorithm}
\usepackage{algorithmic}

\usepackage{newfloat}
\usepackage{listings}
\DeclareCaptionStyle{ruled}{labelfont=normalfont,labelsep=colon,strut=off} 
\floatstyle{ruled}
\newfloat{listing}{tb}{lst}{}
\floatname{listing}{Listing}
\usepackage{tikz}
\usepackage{comment}
\usepackage{amsmath,amssymb} 
\usepackage{color}
\usepackage{booktabs}
\usepackage{epsfig}
\usepackage{paralist}
\usepackage{balance}
\usepackage{float}
\usepackage{multirow}
\usepackage{amsfonts}
\usepackage{cite}

\DeclareMathOperator{\EM}{E_{Pose}} 
\DeclareMathOperator{\D}{D_{Pose}}

\def\ie{\emph{i.e.}}
\def\eg{\emph{e.g.}}

 \nocopyright

\title{ViA: View-invariant Skeleton Action Representation Learning via \\Motion Retargeting}

\author{
    Di Yang\textsuperscript{\rm 1,2}\hskip 1em
    Yaohui Wang\textsuperscript{\rm 1,2,4\thanks{Work done while the author was at Inria}}\hskip 1em
    Antitza Dantcheva\textsuperscript{\rm 1,2}\hskip 1em
    Lorenzo Garattoni\textsuperscript{\rm 3}\\
    Gianpiero Francesca\textsuperscript{\rm 3}\hskip 1em
    François Brémond\textsuperscript{\rm 1,2}\hskip 1em
}
\affiliations{
    \textsuperscript{\rm 1}Inria \hskip 1em 
    \textsuperscript{\rm 2}Université Côte d'Azur \hskip 1em 
    \textsuperscript{\rm 3}Toyota Motor Europe\hskip 1em 
    \textsuperscript{\rm 4}Shanghai AI Laboratory\\
    
{\tt\small \{di.yang, yaohui.wang, antitza.dantcheva, francois.bremond\}@inria.fr} \hskip 1em

{\tt\small \{lorenzo.garattoni, gianpiero.francesca\}@toyota-europe.com}

}

\usepackage{bibentry}

\begin{document}

\maketitle

\begin{abstract}

Current self-supervised approaches for skeleton action representation learning often focus on constrained scenarios, where videos and skeleton data are recorded in laboratory settings. When dealing with estimated skeleton data in \textit{real-world videos}, such methods perform poorly due to the large variations across subjects and camera viewpoints. To address this issue, we introduce ViA, a novel View-Invariant Autoencoder for self-supervised skeleton action representation learning\footnote{Project website: \url{https://walker-a11y.github.io/ViA-project}}. ViA leverages motion retargeting between different human performers as a pretext task, in order to disentangle the latent action-specific `Motion' features on top of the visual representation of a 2D or 3D skeleton sequence. Such `Motion' features are invariant to skeleton geometry and camera view and allow ViA to facilitate both, cross-subject and cross-view action classification tasks. We conduct a study focusing on transfer-learning for skeleton-based action recognition with self-supervised pre-training on real-world data (\eg, Posetics). Our results showcase that skeleton representations learned from ViA are generic enough to improve upon state-of-the-art action classification accuracy, not only on 3D laboratory datasets such as NTU-RGB+D 60 and NTU-RGB+D 120, but also on real-world datasets where only 2D data are accurately estimated, \eg, Toyota Smarthome, UAV-Human and Penn Action.

\end{abstract}

\section{Introduction}
\begin{figure*}
\begin{center}

\includegraphics[width=0.9\linewidth]{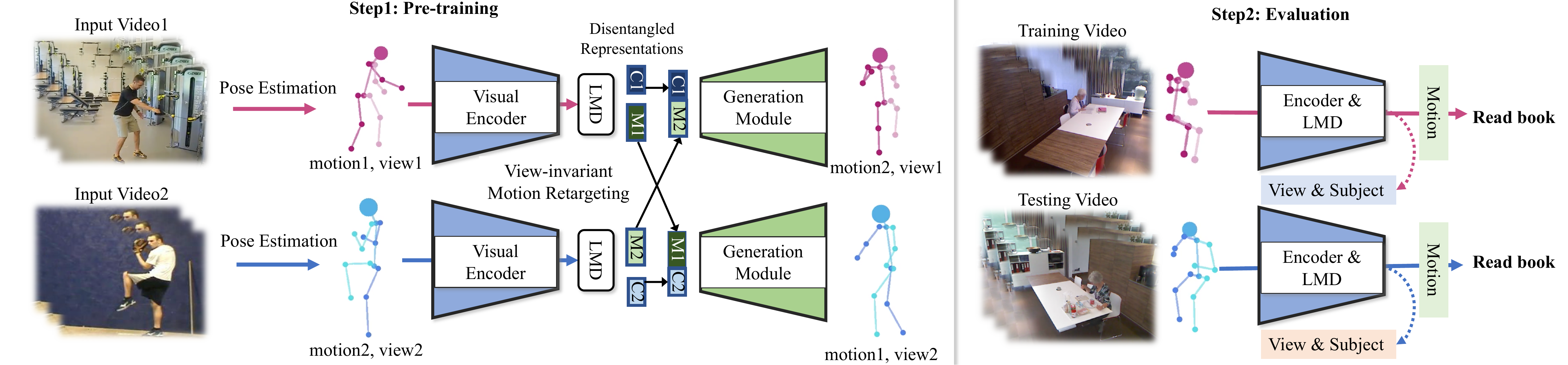}
\end{center}
   \vspace{-0.5cm}
   \caption{ \textbf{General pipeline of ViA.} Our framework consists of two steps. Firstly, we apply view-invariant motion retargeting as a pretext task for pre-training a view- and subject-invariant autoencoder for given skeleton sequences. Secondly, the pre-trained visual encoder is evaluated by transferring to downstream cross-view and cross-subject action classification tasks. LMD: Latent Motion Disentanglement module. M: `Motion' features, C: `Character' features.}
\vspace{-0.4cm}
\label{fig:intro}
\end{figure*}

Human action recognition is a crucial task in real-world video understanding. Recent works have made promising progress by adopting spatio-temporal Convolutional Neural Networks (CNNs)~\cite{3dcnn, Carreira_2017_CVPR, 3d-resnet, slow-fast, x3d, Ryoo2020AssembleNetAM, li2021ctnet, Wang_2021_CVPR} or Transformer~\cite{arnab2021vivit} to effectively extract features from RGB videos and optical flows~\cite{two-stream, c-2stream}. 
However, such methods are generally poor at recognizing actions performed by subjects or recorded from viewpoints that are not present in the training data. One of the open challenges is to design methods that are invariant to subject and view. As an alternative, skeleton-based action recognition methods have been demonstrated robust to changes in viewpoints and subjects~\cite{Vemulapalli2014HumanAR, Caetano2019SkeletonIR, Yan2018SpatialTG, Li_2019_CVPR, 2sagcn2019cvpr, res-gcn, topology_2021_ICCV, duan2021revisiting} when the models were trained (and evaluated) on laboratory indoor simulation datasets~\cite{Shahroudy2016NTURA, NTU-120, ucla_2014_CVPR}, where high-quality 3D skeleton data are accessible. 
Unfortunately, 3D skeleton-based models often fail to generalize to real-world videos~\cite{Carreira_2017_CVPR, penn, Das_2019_ICCV, uav} containing realistic and diverse conditions, whereas acquiring high-quality 3D skeleton labeled data is extremely expensive.
On the contrary, many studies~\cite{weinzaepfel2021mimetics, unik, duan2021revisiting} have shown that 2D estimated skeleton, although sensitive to view and subject variations, is more accurate and more effective for action recognition compared to estimated 3D counterparts in many real-world scenarios~\cite{penn, Das_2019_ICCV, uav, Carreira_2017_CVPR}. Based on this observation, we hypothesize that action recognition, particularly based on 2D skeletons, could be improved by embedding a view-invariant representation of skeleton sequences. 

In this work we propose ViA, 
a novel framework to train an embedding model that represents a view-invariant 2D action without the need to explicitly reconstruct 3D skeletons or camera parameters. Given a pair of videos, we denote the temporal static information of a skeleton sequence, \ie, `viewpoints', `body size', etc., as `Character', while the temporal dynamic information, \ie, the specifics of the `action' performed by the subject, as `Motion'. ViA is able to make subjects perform each other's action, while maintaining viewpoint and body size invariance. As the learned `Motion' representation is subject and view agnostic, it can be effectively applied for cross-subject and cross-view action recognition. 
Specifically, as shown in Fig.~\ref{fig:intro}, ViA incorporates 
\begin{inparaenum}[(i)]
\item an encoder that operates the skeleton sequences extracted from source and driving videos, along with 
\item a Latent Motion Disentanglement module (LMD) to learn the `Motion' and `Character' components by orthogonal decomposition from the visual skeleton representations. 
\item A lightweight skeleton sequence decoder is employed to generate the novel pair of skeleton sequences from the representations where their `Motion' features are swapped with each other. 
\end{inparaenum}
In addition, towards independence from action annotations, we propose to train the autoencoder of ViA in a self-supervised manner by adopting contrastive loss and a cycle of reconstruction loss.

Towards assessing the performance of our framework, we first pre-train ViA on the large-scale real-world Posetics dataset with a rich variety of subjects and viewpoints and we evaluate the quality of the learned action representation by fine-tuning and linear evaluation protocols on unseen 2D real-world action recognition datasets (\eg, Toyota Smarthome, UVA-Human and Penn Action). As ViA is not limited to 2D skeletons, we additionally validate the effectiveness of ViA on laboratory 3D datasets (\eg, NTU-RGB+D 60 and 120). Experimental analyses confirm that through motion retargeting, ViA outperforms state-of-the-art methods~\cite{sun2020viewinvariant, ConNTU, orvpe, motionconsistency} on self-supervised action representation learning and the learned video representations can notably transfer to videos with cross-view and cross-subject challenges. 

In summary, the contributions of this paper are the following.
\begin{inparaenum}[(i)]
\item We introduce ViA, a novel joint generative and discriminative framework. ViA leverages motion retargeting as a pretext task to learn view- and subject-invariant skeleton-based action representations.
\item We introduce a novel Latent Motion Disentanglement mechanism to decompose and regroup the view-agnostic `Motion' features for skeleton sequences by orthogonal decomposition on visual representations in the latent space. 
\item We set a new state-of-the-art for self-supervised skeleton action recognition on the real-world Posetics dataset. 
\item We conduct a study and show that pre-training ViA on Posetics and transferring it onto an unseen target dataset represents a generic and effective methodology for view- and subject-invariant action classification.

\end{inparaenum}

\section{Related Work}
\begin{figure*}[t]
\begin{center}

\includegraphics[width=0.89\linewidth]{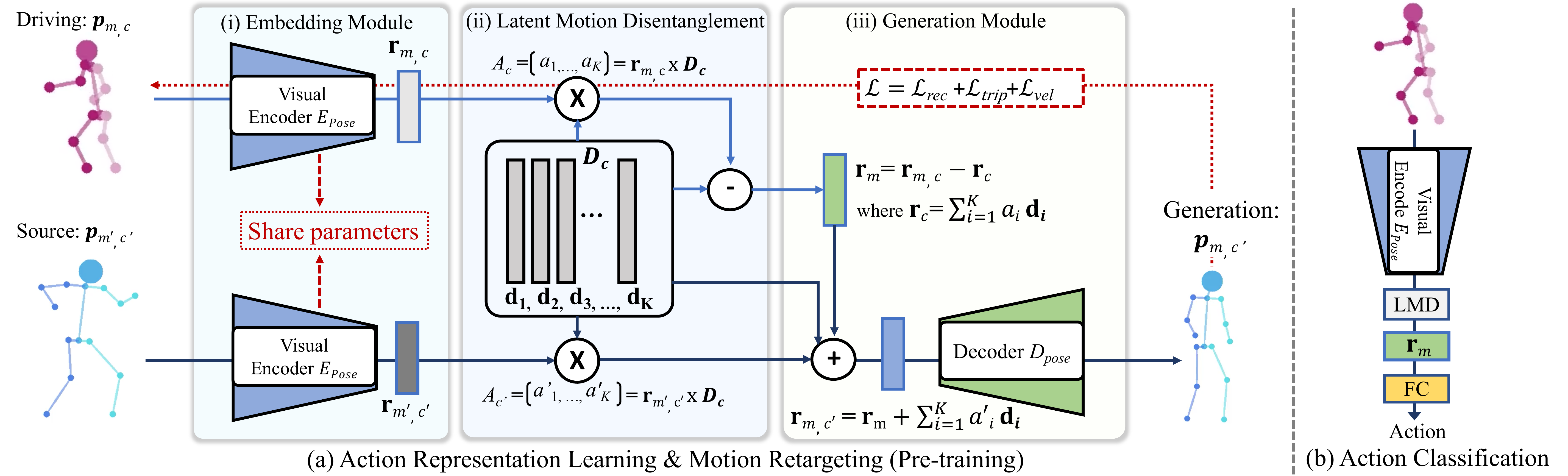}
\end{center}
   \vspace{-0.5cm}
   \caption{\textbf{Overview of the proposed framework.} ViA is an autoencoder consisting of two networks, a visual encoder $\EM$ and a decoder $\D$.  In the latent space, we apply Latent Motion Disentanglement (LMD) by learning a character dictionary $\mathbf{D}_c$, which is an orthogonal basis where each vector represents a basic visual transformation of the `Character' aspects. (a) In the \textbf{pre-training} stage, ViA takes two skeleton sequences $\mathbf{p}_{m,c}$ and $\mathbf{p}_{m',c'}$ as driving and source sequence respectively. Firstly, the two samples are encoded into latent codes $\mathbf{r}_{m, c}$ and $\mathbf{r}_{m',c'}$. Then, their projections $A_c$ and $A_{c'}$ along $\mathbf{D}_c$ can be computed and the linear combination of $A_c$ (or $A_{c'}$) with $\mathbf{D}_c$ is considered as the `Character' features. The `Motion' features of the driving sequence $\mathbf{r}_{m}$ can be disentangled and regrouped with the `Character' features of $\mathbf{p}_{m',c'}$ to become the target code $\mathbf{r}_{m,c'}$. Finally, the target skeleton sequence $\mathbf{p}_{m,c'}$ is generated from $\mathbf{r}_{m,c'}$ by $\D$. (b) \textbf{In the action classification} stage, we transfer the pre-trained visual Encoder $\EM$ stacked with LMD onto downstream tasks for fine-tuning or feature extraction.}
\vspace{-0.3cm}
\label{fig:overview}
\end{figure*}

\paragraph{Self-supervised Skeleton Action Representation.} Self-supervised skeleton representation learning involves learning spatio-temporal features from numerous unlabeled data by means of pretext tasks, \eg, motion consistency learning~\cite{motionconsistency} and skeleton colorization~\cite{colorization} for specific skeleton sequence processing. These methods highly rely on the quality of pretext tasks. Recently, contrastive learning and its variants~\cite{non-para, hjelm2019learning, bachman2019learning, tian2020cmc, He_2020_CVPR, chen2020simclr, jiao2020subgraph, orvpe} have established themselves as an important direction for self-supervised representation learning, due to their promising performances. Inspired by~\cite{tian2020cmc}, CrosSCLR~\cite{ConNTU} trains action models~\cite{Yan2018SpatialTG} in three modalities including joint, bone, and motion by encouraging cross-view consistency. Recent techniques~\cite{motionconsistency, colorization, ConNTU} have only shown promising results on laboratory datasets based on high-quality 3D skeleton sequences that are more robust to view variations. They thus struggle to deal with a large diversity of subjects and viewpoints when generalizing onto real-world action recognition tasks especially on 2D skeleton datasets~\cite{Das_2019_ICCV, uav, penn}. 

\vspace{-0.15cm}
\paragraph{View-invariant Skeleton Representation.}
To explore the view-invariant representation ability of human skeletons, previous methods~\cite{view-nips2018, encoderwacv, nie2020unsupervised, nie2021view} aim at disentangling the view-dependent and pose-dependent features by two independent encoders on top of a single 3D skeleton using probabilistic embedding for view-invariant action recognition. To further address inherent ambiguities in 2D skeleton due to 3D-to-2D projection for action recognition, recent methods~\cite{sun2020viewinvariant, cv-mim, bmvc2021unsupervised} perform the disentanglement learning on specific sensors (\eg, motion capture system) capturing multi-view 2D skeletons. However, The aforementioned methods all process the skeleton sequence frame by frame, which are challenged in capturing the temporal features of the sequence and they are often not available when applying for common 2D datasets~\cite{Das_2019_ICCV, uav, penn} where collecting data in multi-view is expensive and challenging.

In our work, ViA applies generative task for the disentanglement and does not need multi-view data and 3D reconstruction. Moreover, unlike previous works only disentangling static aspects `view' and `pose' for a single frame, the disentanglement of ViA is designed for `Character' (including `view' and `pose') and `Motion' coded in a sequence. The important temporal dimension is considered to better generalize to action understanding tasks. By disentangling `Motion' features using orthogonal decomposition in the latent space, ViA eliminates the requirement of explicit regularization terms~\cite{sun2020viewinvariant, cv-mim} that encourage disentanglement and smoothness of the learned representation. 

\vspace{-0.15cm}

\paragraph{Skeleton Generation by Motion Retargeting.}
Motion retargeting is one of the important applications of video generation~\cite{Tulyakov:2018:MoCoGAN,Wang_2020_CVPR,WANG_2020_WACV,yu2022digan,stylegan_v}.
Recent methods have used both spatially structured representations (\eg, 2D skeleton sequences~\cite{NKN_2018_CVPR, 2dmr, dance}) and non-structure representations (\eg, latent codes~\cite{wang2022latent}) to transfer motion between humans. 
Current skeleton generation method~\cite{2dmr} shows that transferring motion across characters can enforce view and motion disentangling.
In this work, we propose to leverage such a generative task to improve action recognition downstream tasks. As opposed to finding directions corresponding to individual visual transformations, we seek to learn a set of directions that cooperatively allows for high-level visual transformations that can be beneficial in skeleton generation. Hence, deviating from existing methods~\cite{NKN_2018_CVPR, 2dmr, dance} using separate networks to learn disentangled features, ViA integrates both, latent motion code, as well as view features in a single encoder, which highly reduces the model complexity and simplifies training. 
In addition, we employ cycle-consistency~\cite{CycleGAN2017, aaai2022remote} to generalize to real-world videos using non-paired skeleton sequences as the only training data. 

\section{Proposed Approach}

In this section, we introduce the full architecture and the training strategy of the ViA framework.

\vspace{-0.1cm}

\paragraph{Overview.} ViA is an autoencoder consisting of a visual encoder and a decoder for skeleton (\ie, pose) sequences (see Fig.~\ref{fig:overview}). We aim at training the visual encoder that can embed the input skeleton sequence to a view- and subject-invariant representation, then transferring the visual encoder for skeleton-based action classification tasks. To disentangle `Motion' from `Character', we apply motion retargeting as the pretext task (\ie, transferring the motion of driving skeleton sequence to the source skeleton sequence maintaining the source skeletons invariant in viewpoint and body size). We note that input skeletons can be in 3D or 2D only. 
The framework is learned with a cycle of reconstruction (the autoencoder is reused taking generated skeleton sequences as the input to recover the original source and driving skeleton sequences) and contrastive losses as self-supervision. We present the architecture details in the following sections.

\subsection{Motion Retargeting Architecture}\label{s3.1}

\paragraph{Skeleton Sequence Modeling.}
The input skeleton sequence with `Character' $c$ and `Motion' $m$ is modeled by a spatio-temporal matrix, noted as $\mathbf{p}_{m,c}\in \mathbb{R}^{T \times V \times C_{in}}$. $T$, $V$, and $C_{in}$ respectively represent the length of the video, the number of body joints in each frame, and the input channels ($C_{in}=2$ for 2D data, or $C_{in}=3$ if we use 3D skeletons). For each frame, the body joint coordinates are arranged in a matrix, with consistent order between different frames. The matrices of different frames are then stacked along the temporal dimension to obtain the matrix $\mathbf{p}_{m,c}$.

\vspace{-0.1cm}

\paragraph{Skeleton Sequence Embedding} \label{sec:embedding}
As shown in Fig.~\ref{fig:overview} a.(i), ViA adopts a visual encoder $\EM$ to extract features of the input driving skeleton sequence $\mathbf{p}_{m,c}$ and represents the features as $\mathbf{r}_{m,c} \in \mathbb{R}^{T' \times C_{out}} = \EM (\mathbf{p}_{m,c})$. $T'$ is the size of temporal dimension after convolutions and $C_{out}$ is the output channel size. In order to encourage the motion retargeting, we select a source skeleton sequence $\mathbf{p}_{m',c'}$ as a contrastive sample. For the selection of this contrasitive sample which is expected to have different motion with driving sequence, we first cluster the skeleton sequences in the training set using K-Means, and then randomly sample a sequence excluding the cluster where $\mathbf{p}_{m,c}$ belongs. The visual representation of the contrastive sample is obtained as $\mathbf{r}_{m',c'}= \EM (\mathbf{p}_{m',c'})$.

Our goal is to learn the view-invariant and generalizable skeleton action representation, which is generated by the visual encoder $\EM$. This encoder thus needs to have a strong capability to extract skeleton spatio-temporal features. To this end, we adopt the recent topology-free skeleton backbone network UNIK~\cite{unik} as the visual encoder. Specifically, 
the $\EM$ is composed of $10$ convolutional building blocks. Each building block contains a spatial network and a temporal convolutional network to extract both spatial and temporal multi-scale features from the skeleton sequence (see Tab.~\ref{tab_arc}). For the spatial processing, we utilize $1 \times 1$ convolutions to expand the data channels and then multiply the features 
by uniformly initialized~\cite{kaiming} and learnable dependency matrices (which replace the adjacency matrices used in GCN-based methods~\cite{Yan2018SpatialTG, 2sagcn2019cvpr, liu2020disentangling, topology_2021_ICCV}). For the temporal processing, we utilize $9 \times 1$ convolutions with strides. The size of the temporal dimension of embedded latent `Motion' $T'$ depends on the duration of the input sequence. For transfer-learning on downstream tasks, we attach the visual encoder to a temporal global average pooling layer and a fully-connected layer followed by a Softmax Layer. 
The output size of each fully-connected layer depends on the number of action classes. Then, we re-train the network with action labels. 

\begin{table}[t]
\centering
\begin{center}
\scalebox{0.9}{

\begin{tabular}{ c| c |c }
\hline
Stages &$\EM$ & $\D$\\

\hline
\hline
\multirow{2}*{Input} & 2D sequence & Rep. \\
& [$T \times V, ~2$] & [$T', ~256$]\\

\hline

\multirow{2}*{1} & \multirow{2}*{
Conv$\left( \begin{array}{cc}
      1\times 1,&64 \\
      9\times 1,&64
\end{array} \right) \times 4$ }

&
Upsample(2)

\\
&&
Conv$\left( \begin{array}{cc}
      7,& 128
\end{array} \right)$ \\
\hline

\multirow{2}*{2} &
\multirow{2}*{Conv$\left( \begin{array}{cc}
    1 \times 1,& 128 \\
    9 \times 1,& 128
     \end{array} 
     \right) \times 3$ }
&Upsample(2)
\\
&& Conv$\left( \begin{array}{cc}
      7,& 64
\end{array} \right)$ \\  

\hline
\multirow{2}*{3}&
\multirow{2}*{Conv$\left( \begin{array}{cc}
    1 \times 1,& 256 \\
    9 \times 1,& 256
     \end{array} 
     \right) \times 3$} 

& Upsample(2)

\\
&& 
Conv$\left( \begin{array}{cc}
      7,& 2V
\end{array} \right)$ \\
\hline
\multirow{1}*{4}&\multirow{1}*{S-GAP $(2 \times V,~256)$ }
&-\\

\hline
\multirow{1}*{Rep.}&[$T', ~256$]&-  \\
\hline
5& T-GAP $(T',~256)$ &-\\
\hline
6& FC, Softmax&-\\
\hline
\multirow{2}*{Output}&\multirow{2}*{Action Class} &2D sequence\\
&&[$T, ~2V$] \\

\hline

\end{tabular}}
\end{center}
\vspace{-0.4cm}
\caption{\textbf{Main building blocks of the Encoder and Decoder networks.} The dimensions of kernels are denoted by $t \times s, c$ (2D kernels) and $t, c$ (1D kernels) for temporal, spatial, channel sizes. S/T-GAP, FC denotes temporal/spatial global average pooling, and fully-connected layer respectively. Rep. indicates the learned representation.}
\vspace{-0.3cm}
\label{tab_arc}
\end{table}  

\vspace{-0.1cm}
\paragraph{Latent Motion Disentanglement.}
Latent Motion Disentanglement (LMD) is the key module of ViA to obtain the `Motion' features on top of the visual representations (see Fig.~\ref{fig:overview} a.(ii)). Our insight is that the `Character' and `Motion' features are independent, these two components can be explicitly discovered in the latent code using the vector orthogonal decomposition method. As the `Character' features are temporal static, we simply represent such features by a one-dimensional vector $\mathbf{r}_c \in \mathbb{R}^{C_{out}}$ for the skeleton sequence $\mathbf{p}_{m,c}$. To model $\mathbf{r}_c$ , we first pre-define a learnable orthogonal basis, noted as $\mathbf{D}_c = \{ \mathbf{d_1}, \mathbf{d_2}, ..., \mathbf{d_K} \}$ with $K \in [1, C_{out})$ where each vector indicates a basic visual transformation of the `Character' aspects of the skeletons. Then, we consider the `Character' features $\mathbf{r}_c$ as a linear combination between this orthogonal basis $\mathbf{D}_c$, and associated magnitudes $A_c=\{ a_1, a_2, ..., a_K \}$. For $\mathbf{p}_{m',c'}$, we can obtain its `Character' component $\mathbf{r}_{c'}$ in the same way:
\begin{equation}\label{chara}
\begin{aligned}
    &\mathbf{r}_{c} = \sum_{i=1}^{K}a_i \mathbf{d_i},&
    &\mathbf{r}_{c'} = \sum_{i=1}^{K}a'_i \mathbf{d_i}.
\end{aligned}
\end{equation}
Subsequently, the `Motion' features $\mathbf{r}_{m}\in \mathbb{R}^{T' \times C_{out}}$ can be obtained by the vector decomposition: $\mathbf{r}_{m}=\mathbf{r}_{m, c}-\mathbf{r}_{c}$. As $\mathbf{r}_{m, c}$ has the temporal dimension of size $T'$, for each feature in the temporal dimension, we repeat the decomposition process with the same `Character' component. According to the Gram-Schmidt algorithm, the set of magnitudes $A_c$ and $A'_c$ can be computed as the projections of $\mathbf{r}_{m, c}$ onto $\mathbf{D}_c$ (see Eq.~\ref{A}) to satisfy the orthogonality of $\mathbf{r}_{m}$ and $\mathbf{r}_{c}$.
\begin{equation}\label{A}
\begin{aligned}
    &a_i = \frac{<\mathbf{r}_{m, c} \cdot \mathbf{d}_i>}{\left\| \mathbf{d}_i\right\| ^2},&
    &a'_i = \frac{<\mathbf{r}_{m', c'} \cdot \mathbf{d}_i>}{\left\| \mathbf{d}_i\right\| ^2}.
\end{aligned}
\end{equation}
With such trained LMD, in the inference stage, we can generate skeleton sequences in multiple viewpoints with a single input skeleton sequence by only changing the magnitudes $A_c$ along the orthogonal basis in the latent space without the need for paired data.

\vspace{-0.1cm}
\paragraph{Skeleton Sequence Generation.}
To generate skeleton sequences, the output $\mathbf{r}_m$ and $\mathbf{r}_{c'}$ are composed and fed into a decoder $\D$ (Fig.~\ref{fig:overview} a.(iii)). 
Unlike $\EM$, the Pose Sequence Decoder $\D$ is lightweight and designed by multiple 1D temporal convolutions with temporal global max pooling and upsampling to respectively encode and decode the skeletons as in~\cite{2dmr}.
The new retargeted skeleton sequence with `Motion' $m$, and `Character' $c'$, noted as $\mathbf{p}_{m,c'}$ is generated from the recombined features, $\mathbf{r}_m+ \mathbf{r}_{c'}$. Similarly, $\mathbf{p}_{m',c}$ can also be generated by swapping the source and driving sequences. The skeleton sequence generation can be formulated as $\mathbf{p}_{m,c'} = \D (\mathbf{r}_m, \mathbf{r}_{c'})$ and $\mathbf{p}_{m',c} = \D (\mathbf{r}_{m'}, \mathbf{r}_c)$.

\subsection{Self-supervised Training}\label{s3.2}
In this section, we introduce the self-supervised training loss function $\mathcal{L}$, which consists of three components, namely reconstruction loss $\mathcal{L}_{rec}$, contrastive triplet loss $\mathcal{L}_{trip}$, and temporal velocity loss $\mathcal{L}_{vel}$:
\begin{equation}\label{l-loss2}
\mathcal{L} = \mathcal{L}_{rec} + \mathcal{L}_{trip} + \mathcal{L}_{vel}.
\end{equation}

\vspace{-0.1cm}
\paragraph{Reconstruction Loss.}
The reconstruction loss aims at guiding the network towards a high generation quality at the global sequence level. It consists of two components: $\mathcal{L}_{rec} = \mathcal{L}_{self} + \mathcal{L}_{cycle}$.
Specifically, at every training iteration, the decoder network $\D$ is firstly required to reconstruct each of the original input samples $\mathbf{p}_{m,c}$ using its representations $\mathbf{r}_{m}$ and $\mathbf{r}_{c}$. This component of the loss is denoted as $\mathcal{L}_{self}$ and formulated as a standard autoencoder reconstruction loss (see Eq.~\ref{recloss1}). Moreover, at each iteration, the decoder is also encouraged to re-compose new combinations. 
To this end, we can explicitly apply the cycle reconstruction loss $\mathcal{L}_{cycle}$ (see Eq.~\ref{recloss1},~\ref{reacloss2}) through the cycle generation. Specifically, the Embedding and LMD modules are used again to disentangle and re-combine the features for the previous generated skeleton sequence $\mathbf{p}_{m',c}$ and $\mathbf{p}_{m,c'}$. Then, the generation module is also used again to reconstruct the original sequence $\mathbf{p}_{m,c}$.
\begin{equation}\label{recloss1}
\begin{aligned}
\mathcal{L}_{self} = \mathbb{E} [\left\| \D (\mathbf{r}_m, \mathbf{r}_c)- \mathbf{p}_{m,c}\right\|^2],\\
\mathcal{L}_{cycle} = \mathbb{E} [\left\| \D (\mathbf{\hat{r}}_m, \mathbf{\hat{r}}_c)- \mathbf{p}_{m,c}\right\|^2],
\end{aligned}
\end{equation}
where
\begin{equation}\label{reacloss2}
\small
\begin{aligned}
 &\mathbf{\hat{r}}_c 
 = \sum_{i=1}^{K} \frac{<\EM (\mathbf{p}_{m',c}) \cdot \mathbf{d}_i>}{\left\| \mathbf{d}_i\right\| ^2} \mathbf{d_i}, &\mathbf{\hat{r}}_{m'} = \EM (\mathbf{p}_{m',c}) - \mathbf{\hat{r}}_c, \\
 &\mathbf{\hat{r}}_{c'} 
 = \sum_{i=1}^{K} \frac{<\EM (\mathbf{p}_{m,c'}) \cdot \mathbf{d}_i>}{\left\| \mathbf{d}_i\right\| ^2} \mathbf{d_i}, &\mathbf{\hat{r}}_{m} = \EM (\mathbf{p}_{m,c'}) - \mathbf{\hat{r}}_{c'}. 
\end{aligned}
\end{equation}

\vspace{-0.1cm}

\paragraph{Contrastive Triplet Loss.}
At the representation level, we adopt a triplet loss $\mathcal{L}_{trip}$ for the sampled skeletons (both driving skeleton sequence and its contrastive sample) as $\mathcal{L}_{trip} = \mathcal{L}_{trip\_M} + \mathcal{L}_{trip\_C}$.
This loss aims to enhance the mutual information of $\mathbf{r}_m$ and $\mathbf{\hat{r}}_m$, which are the representations of the same `Motion' performed by different characters, to produce view-invariant `Motion' representations. Specifically, we encourage the similarity between $\mathbf{r}_m$ and $\mathbf{\hat{r}}_m$, while discouraging the similarity between $\mathbf{\hat{r}}_m$ and the other `Motion' performed in its contrastive sample, $\mathbf{r}_{m'}$. Similarly, we define the `Character' triplet loss in the same way. The loss components are described as follows (the triplet margin $\alpha=1.0$):
\begin{equation}\label{triplossM}
\begin{aligned}
    &\mathcal{L}_{trip\_M} = \mathbb{E} [\left\| \mathbf{\hat{r}}_m -\mathbf{r}_m \right\| -
    \left\|  \mathbf{\hat{r}}_m -\mathbf{r}_{m'}\right\| + \alpha]_+, \\
    &\mathcal{L}_{trip\_C} = \mathbb{E} [\left\| \mathbf{\hat{r}}_{c} -\mathbf{r}_{c} \right\| - 
    \left\| \mathbf{\hat{r}}_{c} -\mathbf{r}_{c'}\right\| + \alpha]_+.
\end{aligned}
\end{equation}

\vspace{-0.1cm}

\paragraph{Velocity Loss.}
As described in~\cite{2dmr}, the use of reconstruction loss only for sequence-level generation produces large errors for end joints such as hands and feet, which gives rise to the foot-skating phenomenon. We argue that the reconstruction loss constrains the network to generate the original skeletons with minimum global errors for all joints, however it misses the important temporal consistencies of each individual joint. We thus explicitly adopt a temporal consistency restriction loss for all $V$ body joints (noted as an ensemble $\mathcal{J}$), which constrains the velocity---\ie, joints shifting along the temporal dimension---of the skeleton sequence (see Eq.~\ref{vel}).
\begin{equation}\label{vel}
\mathcal{L}_{vel} = \lambda \mathbb{E} [ \sum_{n\in \mathcal{J}} \left\| \mathcal{V}_{n}\big(\D (\mathbf{\hat{r}}_m, \mathbf{\hat{r}}_c)\big) - \mathcal{V}_{n}(\mathbf{p}_{m,c}) \right\| ^2],
\end{equation}
where $\mathcal{V}_n$ denotes the velocity of the $n$-th joint, which can be calculated by the distance between this skeleton joint at frame $\tau$-th and at frame $\tau+1$-th. $\lambda$ indicates the weighting factor of the velocity loss.

\subsection{Transfer-learning for Action Classification}
ViA aims at pre-training a generic and view-invariant visual encoder. The model properties are verified by transfer-learning of $\EM$ (stacked with LMD) for action recognition tasks (see Fig.~\ref{fig:overview} (b)). In practice, we attach the visual encoder, where the pre-trained weights are used as initialization, to a temporal global average pooling layer and a fully-connected layer followed by a Softmax Layer. 
The output size of each fully-connected layer depends on the number of action classes. Then, we re-train the network with action labels 
on the target datasets.
Following common evaluation protocols used in previous unsupervised action representation frameworks~\cite{ConNTU, colorization, motionconsistency, orvpe}, we conduct both \textit{linear evaluation} by training only the fully-connected layer with the backbone frozen, and \textit{fine-tuning evaluation} by further refining the whole network on downstream tasks.

\section{Experiments and Analysis}\label{sec:exp}

We conduct extensive experiments to evaluate ViA. 
Firstly, 
we compare ViA 
against the state-of-the-art self-supervised models on the large-scale pre-training dataset \textbf{Posetics}, by linear evaluation 
Secondly, we study the generalizability of ViA to 
quantify the performance improvement obtained by transfer-learning on the target 2D datasets (\ie, \textbf{Toyota Smarthome, UAV-Human} and \textbf{Penn Action}) as well as 3D datasets (\ie, \textbf{NTU-RGB+D 60} and \textbf{NTU-RGB+D 120}) after pre-training on Posetics. 
Thirdly, we evaluate the quality of the motion (\ie, action) generated by the retargeting module on the synthetic dataset \textbf{Mixamo}.
Finally, we provide 
an exhaustive ablation study of ViA.

\subsection{Datasets and Evaluation Protocols}

\noindent\textbf{Posetics}~\cite{unik} is created on top of Kinetics-400~\cite{Carreira_2017_CVPR} videos. It contains 142,000 real-world video clips of 320 action classes with the corresponding 2D and 3D skeletons. We use the Posetics dataset to pre-train our action representation learning framework with skeleton data and we study the transfer-learning on skeleton-based action classification. We use Top-1 and Top-5 accuracy as evaluation metrics~\cite{unik}.
\vspace{0.05cm}

\noindent\textbf{Toyota Smarthome}~\cite{Das_2019_ICCV} (Smarthome) is a real-world dataset for daily living action classification and contains 16,115 videos of 31 action classes. It provides RGB videos, 2D and 3D skeleton data~\cite{Yang_2021_WACV}. As the provided 2D data is more robust for action recognition even for cross-view evaluation~\cite{Yang_2021_WACV, unik}, unless otherwise stated, we use 2D data for the experiments. For the evaluation, we report mean per-class accuracy following the cross-subject (CS) and cross-view (CV1 and CV2) evaluation protocols.

\vspace{0.05cm}

\noindent\textbf{UAV-Human}~\cite{uav} contains 22,476 video sequences collected by a flying UAV including 2D skeleton data estimated by~\cite{fang2017rmpe}. In this work, we use only 2D skeleton data and we follow Cross-subject (CS1 and CS2) evaluation protocols.

\vspace{0.05cm}

\noindent\textbf{Penn Action}~\cite{penn} contains 2,326 video sequences of 15 different actions. In this work, we use 2D skeletons obtained by LCRNet++~\cite{RogezWS18} for experiments and we report Top-1 accuracy following the standard train-test split.

\vspace{0.05cm}

\noindent\textbf{NTU-RGB+D 60}~\cite{Shahroudy2016NTURA} consists of 56,880 sequences of high-quality 3D skeletons, captured by the Microsoft Kinect v2 sensor. We only use sequences of 3D skeletons in this work and we follow the cross-subject (CS) and cross-view (CV) evaluation protocol.

\vspace{0.05cm}

\noindent\textbf{NTU-RGB+D 120}~\cite{NTU-120} extends the number of action classes and videos of NTU-RGB+D 60 to 120 classes 114,480 videos. We use 3D skeleton sequences and we follow the cross-subject (CS) and cross-set (CSet) evaluation protocols. 
\vspace{0.05cm}

\noindent\textbf{Mixamo}~\cite{mixamo} is a 3D animation collection, which contains approximately 2,400 unique motion sequences, including elementary actions, and various dancing moves. Each of these motions may be applied to 71 distinct characters, which share a human skeleton topology, but may differ in their body size and proportions. We use such a synthetic dataset, which includes the cross-reconstruction ground truth (\ie, the same motion pattern performed by different characters and in different viewpoints obtained by rotated 3D and projected 2D skeletons) for pre-training and evaluating the motion retargeting module in ViA.

\subsection{Evaluation on Self-supervised Action Classification}\label{sec:posetics}
Our objective is to improve action recognition performance on 2D skeleton datasets by learning an action representation on a sufficiently large dataset. Hence, 
in this section, we evaluate ViA on self-supervised action classification (\ie, linear evaluation) on the large-scale \text{Posetics} dataset and then compare ViA with state-of-the-art approaches.

\vspace{-0.1cm}

\paragraph{Comparison with State-of-the-art (SoTA).}
For fair comparison, we re-implement recent state-of-the-art skeleton-based action representation learning approaches~\cite{sun2020viewinvariant, ConNTU, orvpe} 
on the Posetics dataset using 2D skeleton data. Results are depicted in Tab.~\ref{tab_posetics} (top): ViA is more effective when compared to 3D-based methods~\cite{ConNTU} applied onto 2D real-world datasets.
Intuitively, we think that
the variation of subject body sizes and of the viewpoints might weaken the robustness of the SoTA embedding networks. In contrast, ViA encourages similarity of the representation for actions performed by different subjects under different viewpoints. This shows that our model is more effective and robust to real-world videos. 
Compared to previous view-invariant embedding approaches~\cite{sun2020viewinvariant} based on single frame, our method considering temporal features is more robust for action recognition. 
In 
Tab.~\ref{tab_posetics} (bottom) we compare fine-tuning results of ViA to other supervised methods~\cite{tcn, Yan2018SpatialTG, 2sagcn2019cvpr, res-gcn, liu2020disentangling, unik} that are trained without representation learning (\ie, training from scratch). Compared to the UNIK backbone model used in our work~\cite{unik}, the pre-training provides minor improvement, as the training data (\ie, Posetics) is sufficiently large. However, when transferring ViA onto smaller benchmark datasets, the impact of representation learning is significant (see Sec.~Evaluation on Transfer-learning).

\begin{table}[t]
\centering
\begin{center}
\scalebox{0.9}{

\setlength{\tabcolsep}{1.6mm}{
\begin{tabular}{ l c c }
\hline
\multirow{2}*{\textbf{Methods}} & \multicolumn{2}{c}{\textbf{Posetics}} \\

& \text{Top-1(\%)} &\text{Top-5(\%)} \\
\hline

\hline
\text{Linear (Baseline)} &\text{8.2}& \text{21.4}\\
\text{Pr-ViPE~\cite{sun2020viewinvariant}} &\text{17.2}& \text{35.3}\\
\text{OR-VPE~\cite{orvpe}}& \text{14.6}& \text{31.2}  \\
\text{3s-CrosSCLR~\cite{ConNTU}} &\text{18.8}& \text{38.1}  \\
\textbf{ViA (Ours)} &  \textbf{20.7}& \textbf{40.1} \\

\hline
TCNs~\cite{tcn}  & 34.0 &57.2\\
ST-GCN~\cite{Yan2018SpatialTG} & 43.3 &67.3\\

2s-AGCN~\cite{2sagcn2019cvpr}  & 47.0 &70.8 \\
Res-GCN~\cite{res-gcn}& 46.7 &70.6\\
MS-G3D Net~\cite{liu2020disentangling} & 47.1 &70.0 \\
\text{UNIK~\cite{unik}}  &\text{47.6} &\text{71.3} \\
\textbf{ViA (Ours ft.)}  &\textbf{48.0} &\textbf{72.6} \\
\hline

\end{tabular}}}
\end{center}
\vspace{-0.4cm}
\caption{Comparison of Top-1 and Top-5 classification accuracy with state-of-the-art \textbf{unsupervised methods (top)} on Posetics. \textbf{Fully-supervised results (bottom)} with fine-tuning (reported as ft.) are also reported for reference.}
\vspace{0.cm}
\label{tab_posetics}
\end{table}

\begin{table}[t]
\centering

\begin{center}
\scalebox{0.9}{
\setlength{\tabcolsep}{.3mm}{
\begin{tabular}{  l c c c c }

\hline
\multirow{2}*{\textbf{Methods}}
& \multicolumn{2}{c}{\textbf{NTU-60}} & \multicolumn{2}{c}{\textbf{NTU-120}}\\
& \text{CS(\%)} &\text{CV(\%)} &\text{CS(\%)}& {\text{ CSet(\%)}}  \\
\hline
\hline
\text{SeBiRe~\cite{nie2020unsupervised}} &\text{-}& \text{79.7} &\text{-} & - \\
\text{CrosSLR~\cite{ConNTU}} &\text{77.8}& \text{83.4} &\text{67.9} & 66.7 \\
\text{Colorization~\cite{colorization}} &\text{75.2}& \text{83.1} &\text{-} & - \\
\textbf{ViA (Ours)} & \textbf{78.1} & \textbf{85.8} &\textbf{69.2} & \textbf{66.9} \\

\hline
\text{W/o pre-training} &\text{86.5}& \text{94.6}&80.1  &84.5  \\
\textbf{Self-supervised pre-training} & \textbf{89.6}& \textbf{96.4}& \textbf{85.0} &\textbf{86.5} \\
\hline
\end{tabular}}}

\end{center}
\vspace{-0.4cm}
\caption{Comparison with previous self-supervised state-of-the-art by \textbf{linear evaluation (top)} on NTU-RGB+D 60 and NTU-RGB+D 120. Transfer learning results by \textbf{fine-tuning (bottom)} are also reported for reference. }
\vspace{-0.3cm}
\label{tab_ntu}
\end{table}

\begin{table*}[t]
\centering

\begin{center}
\scalebox{0.9}{
\setlength{\tabcolsep}{1.6mm}{
\begin{tabular}{  l c c c c c c c c c c }

\hline
\multirow{2}*{\textbf{Methods}}
& \multirow{2}*{\textbf{Pre-training}}& \multicolumn{4}{c}{\textbf{Toyota Smarthome}} & \multicolumn{3}{c}{\textbf{UAV-Human}}& \multicolumn{2}{c}{\textbf{Penn Action}}\\
& &\text{\#Params}& \text{CS(\%)} &\text{CV1(\%)} &\text{CV2(\%)}&\text{\#Params}& {\text{ CS1(\%)}} & {\text{CS2(\%)}} &\text{\#Params}& {\text{Top-1(\%)}} \\
\hline
\hline
\text{Random init.}&Scratch &7.97K&\text{24.6}& \text{17.2}& 20.7 & 39.85K&3.8  &4.1 & 3.85K &29.8 \\
\text{Supervised}& Posetics w/ labels& 7.97K &\text{51.9}& \text{35.4}& \text{52.2} &39.85K &32.9 & 56.1 &3.85K& \text{97.3} \\
\textbf{Self-supervised}& Posetics w/o labels&7.97K&\textbf{49.5}& \textbf{33.6}& \textbf{52.6} &39.85K &\textbf{29.5} &\textbf{46.7} &3.85K &\textbf{90.2} \\

\hline
Random init. & Scratch&3.45M &63.1 &22.9&61.2 & 3.45M & 39.2 &67.3 &3.45M & 94.0 \\

\text{Supervised}& Posetics w/ labels &3.45M&\text{64.5}& \text{36.1}&\text{65.2} & 3.45M& \text{42.6} &69.5 &3.45M &\text{98.0}\\
\textbf{Self-supervised}& Posetics w/o labels &3.45M&\textbf{64.0}& \textbf{35.6}&\textbf{{65.4}} & 3.45M &\textbf{41.3} &\textbf{68.5} &3.45M &\textbf{97.7} \\

\hline
\multirow{2}*{\text{Previous SoTA}}&-&\multicolumn{4}{l}{\cite{unik}}& \multicolumn{3}{l}{\cite{shift_2020_CVPR}}&   \multicolumn{2}{l}{\cite{sun2020viewinvariant}} \\
 &-&- &\text{63.1}& \text{22.9}&\text{61.2} & - &\text{38.0}& 67.0 &- &\text{97.5} \\
\textbf{ViA (Ours)}& - &-&\textbf{64.5}& \textbf{36.1}&\textbf{65.4} & - &\textbf{42.6}&\textbf{69.5} &- &\textbf{98.0} \\

\hline
\end{tabular}}
}
\end{center}
\vspace{-0.4cm}
\caption{Transfer learning results by \textbf{linear evaluation (top)} and \textbf{fine-tuning (middle)} on Smarthome, UAV-Human and Penn Action with self-supervised pre-training on Posetics compared to Baseline (random initialization). Results with supervised pre-training and \textbf{previous state-of-the-art (bottom)} are also reported.}
\vspace{-0.cm}
\label{tab_tr}
\end{table*}

\begin{table*}[t]
\centering

\begin{center}
\scalebox{0.9}{
\setlength{\tabcolsep}{.8mm}{
\begin{tabular}{  l c c c c c c c c }

\hline
\multirow{2}*{\textbf{Methods}}&\multirow{2}*{\textbf{Pre-training}} & \multirow{2}*{\textbf{Training data}}
  & \multicolumn{3}{c}{\textbf{Toyota Smarthome}} & \multicolumn{2}{c}{\textbf{UAV-Human}}& \multicolumn{1}{c}{\textbf{Penn Action}}\\

& & & \text{CS(\%)} &\text{CV1(\%)} &\text{CV2(\%)}& {\text{ CS1(\%)}} & {\text{CS2(\%)}} & {\text{Top-1 Accuracy(\%)}} \\
\hline
\hline

\text{Random init.~\cite{unik}}&Scratch  & 5\% &\text{22.9}& \text{5.6}& 33.7 &10.9 & 10.4 &32.4 \\

\textbf{Self-supervised}& Posetics w/o labels &  5\% &\textbf{38.6}& \textbf{16.8}& \textbf{42.6} &\textbf{21.7} &\textbf{33.3} & \textbf{65.8} \\

\hline
\text{Random init.~\cite{unik}}& Scratch&  10\% &\text{33.8}& \text{8.5}& 39.5 &17.8 & 25.6 &39.8 \\
\textbf{Self-supervised}& Posetics w/o labels & 10\% &\textbf{45.3}& \textbf{22.7}& \textbf{46.6} &\textbf{31.0} &\textbf{43.7} & \textbf{85.2} \\

\hline
\end{tabular}}}

\end{center}
\vspace{-0.4cm}
\caption{Transfer learning results by \textbf{fine-tuning} on all benchmarks of Smarthome, UAV-Human and Penn Action with randomly selected \textbf{5\% (top)} and \textbf{10\% (bottom)} of labeled training data. }
\vspace{-0.3cm}
\label{tab_fewer}
\end{table*}

\subsection{Evaluation on Transfer-learning}\label{sec:tranfer}
In this section, we study the transfer ability of ViA by both linear evaluation and fine-tuning evaluation with self-supervised training on Posetics. We transfer the model onto three 2D skeleton action classification benchmarks \ie, \text{Toyota Smarthome, UAV-Human} and \text{Penn Action} with no additional pre-training. As Smarthome and UAV-Human mainly focus on the cross-subject and cross-view challenges, the results measure the view-
and subject-invariance of the 2D action representation of ViA models. We also report the results with supervised pre-training for reference (\ie, we add a classifier at the end of the visual encoder and adopt cross entropy loss using action labels during training).
\vspace{-0.1cm}

\paragraph{Linear Evaluation.}
Tab.~\ref{tab_tr} (top) shows the linear results on the three datasets. This evaluates the effectiveness of transfer-learning with fewer parameters (only the classifier is trained) compared to classification from random initialization. The results suggest that the weights of the model can be well pre-trained without action labels, providing a strong transfer ability especially on smaller benchmarks (\eg, +31.9\% Smarthome on CV2 and +70.4\% on Penn Action) and the pre-trained visual encoder is generic enough to extract meaningful action features from skeleton sequences. 

\vspace{-0.1cm}
\paragraph{Fine-tuning.}
Tab.~\ref{tab_tr} (middle) shows the fine-tuning results, when the whole network is re-trained. These results suggest that pre-training can improve upon previous SoTA~\cite{unik} with no pre-training. The self-supervised pre-trained model also performs competitively compared to supervised pre-trained models. From these results we conclude that collecting a large-scale video dataset, without the need of action annotation, can be beneficial to downstream tasks, especially when using our proposed view-invariant ViA for the 2D action classification task (\eg, +12.7\% on Smarthome CV1 and +4.2\% on CV2). Furthermore, we compare our fine-tuning results to other SoTA skeleton-based supervised approaches~\cite{unik, shift_2020_CVPR, sun2020viewinvariant}. The results in Tab.~\ref{tab_tr} (bottom) show that ViA outperforms all previous approaches on all the three real-world datasets. 

\vspace{-0.1cm}

\paragraph{Training with Fewer Labels.}
In some real-world applications, labeled data may be lacking, which makes it challenging to train models with good performance. To evaluate ViA in such cases, we pre-train with Posetics and then fine-tune the visual encoder with 5\% and 10\% of the labeled data. As shown in Tab.~\ref{tab_fewer}, without pre-training, the accuracy of the baseline~\cite{unik}
significantly decreases with the amount of training data. In contrast, ViA still achieves good performance on all three datasets.

\vspace{-0.1cm}
\paragraph{3D Skeleton Action Classification.}\label{sec:tranfer3d} 
As ViA can be simply extended to take 3D skeleton sequence as input, we further analyze the transfer ability of ViA onto 3D skeleton action recognition tasks. We firstly compare~\cite{ConNTU},~\cite{orvpe} and ViA on Posetics using officially provided 3D data (we get \text{17.1\%}, \text{12.9\%} and \text{19.3\%}, respectively for linear evaluation). These results are lower than related results in Tab.~\ref{tab_posetics} using 2D data. We argue that, although 3D skeletons are more robust to the view variation, 2D skeletons extracted from images or videos tend to be more accurate than extracted 3D skeletons. In contrast, laboratory datasets (\eg, \text{NTU-RGB+D 60 \& 120}) provide 3D skeleton data obtained by RGBD sensors that have a higher quality than the one provided by 2D data. To study the impact of action representation learning, we also transfer the ViA pre-trained on Posetics (3D skeletons) without action labels onto NTU-RGB+D-60 and NTU-RGB+D-120 by fine-tuning. The action recognition performance can still be improved (\eg, +4.9\% on NTU-RGB+D-120 CS, see Tab.~\ref{tab_ntu} (bottom)). To compare with other recent self-supervised methods~\cite{nie2020unsupervised, ConNTU, colorization}, we follow the same pre-training setting and linear evaluation protocol and report state-of-the-art accuracy (see Tab.~\ref{tab_ntu} (top)).  
\begin{table}[t]
\centering

\begin{center}
\scalebox{0.9}{
\setlength{\tabcolsep}{1.mm}{
\begin{tabular}{ l c c }
\hline
\textbf{Methods}& \textbf{Sup.}& \textbf{Unsup.}\\

\hline
\hline
\text{NKN~\cite{NKN_2018_CVPR}} &\text{1.51}& \textbf{-}  \\
\text{MotionRetargeting2D~\cite{2dmr}} &\text{0.96} & \text{2.56}  \\
\hline

\text{ViA w/o $\mathbf{D}_c$ (Ours)}  & \text{2.42}& \text{-} \\
\text{ViA w/ $\mathbf{D}_c$ (Ours)}  \\
\text{~~~~size ${K}= 2$}  & \text{1.16}& \text{-} \\
\text{~~~~size ${K}= 64$}  & \text{0.89}& \text{-} \\
~~~~size \textbf{${K}= 32$}  & \textbf{0.86}& \textbf{2.47} \\
\hline
\end{tabular}}}
\end{center}
\vspace{-0.4cm}
\caption{Quantitative comparisons of Mean Square Error (MSE) show that our framework outperforms other SoTA motion retargeting methods on Mixamo.}
\vspace{-0.3cm}
\label{tab_mr}
\end{table}

\subsection{Evaluation on Motion Retargeting}\label{sec:retargeting}
As motion retargeting is our pretext task, we here evaluate the proposed LMD mechanim of ViA by motion retargeting performance.
We randomly split training and test sets on the \text{Mixamo dataset} and follow the same setting and protocol described in~\cite{2dmr}. As previous SoTA~\cite{NKN_2018_CVPR, 2dmr} are supervised approaches, for fair comparison, we also train ViA using cross reconstruction loss on~\cite{mixamo} in a supervised manner. To validate the impact of proposed $\mathbf{D}_c$, we learn $\mathbf{r}_{c}$ by decomposing $\mathbf{r}_{m, c}$ on a pre-defined and fixed subspace without the learnable $\mathbf{D}_c$. From the evaluation results reported in Tab.~\ref{tab_mr}, we observe that in the absence of $\mathbf{D}_c$, model fails to generate high-quality skeletons, which proves the effectiveness of $\mathbf{D}_c$. Then we conduct an ablation analysis on the size of $\mathbf{D}_c$. The results suggest that motion retargeting by ViA (w/ $\mathbf{D}_c$ in including $32$ directions) performs the best and achieve SoTA accuracy. We also report unsupervised results by cycle consistency learning.

\vspace{-0.1cm}

\begin{figure}[t]
\begin{center}

\includegraphics[width=1\linewidth]{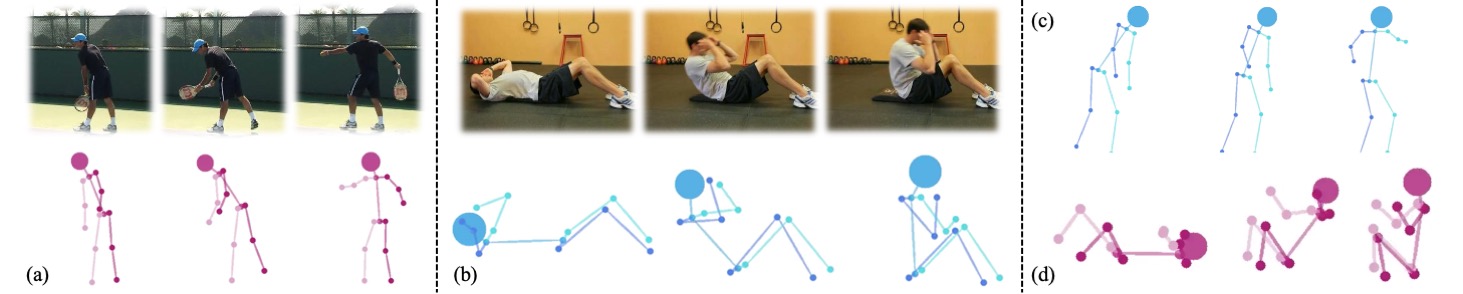}
\end{center}
   \vspace{-0.3cm}
   \caption{Qualitative results on \textbf{Motion Retargeting.} (a) and (b) are the input pair of videos and corresponding 2D skeleton sequences. (c) is the generated 2D skeleton sequence that represents the character of (b) performing the motion in (a) while maintaining the viewpoint and body size invariance. (d) is the generated 2D skeleton sequence that represents the character of (a) performing the motion in (b).}
\vspace{0.cm}
\label{fig:gen}
\end{figure}

\begin{figure}[t]
\begin{center}

\includegraphics[width=1\linewidth]{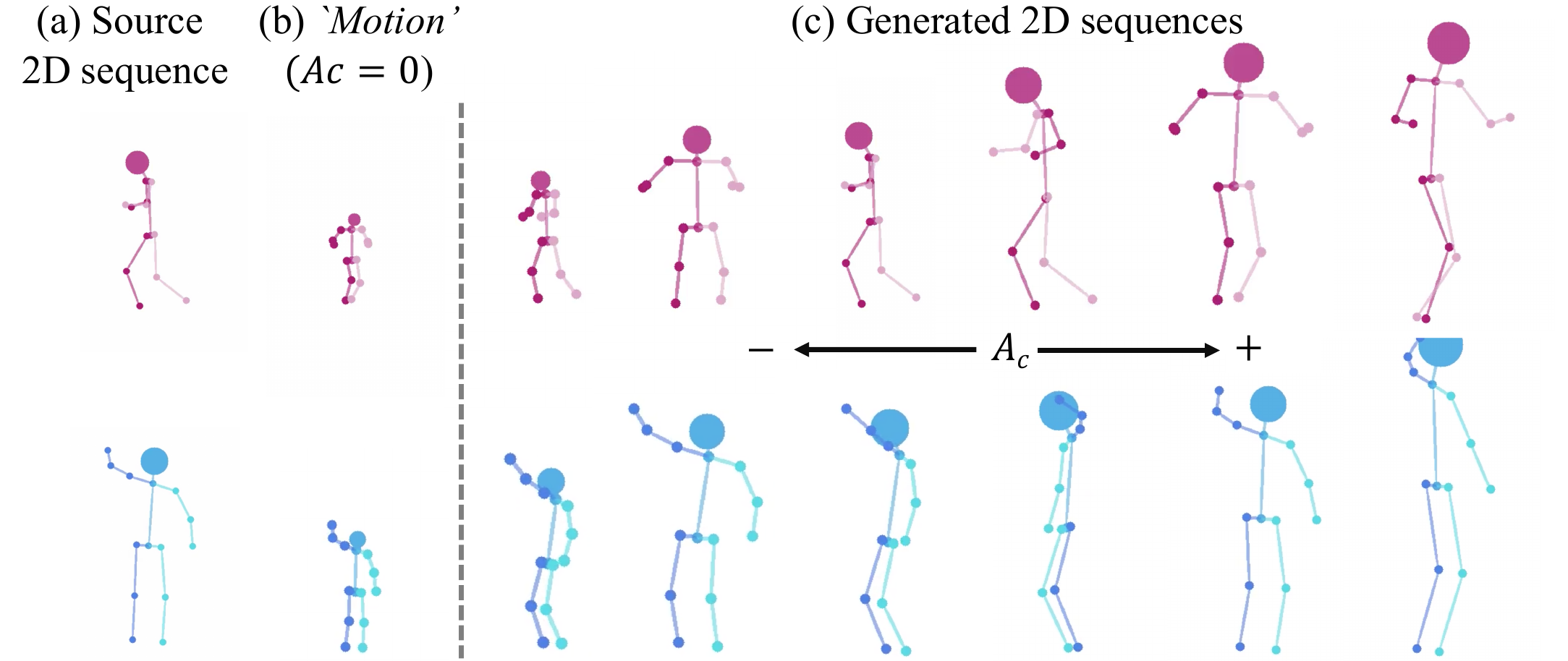}
\end{center}
   \vspace{-0.35cm}
   \caption{Qualitative results on \textbf{2D Motion Generation.} Given a source skeleton sequence, we can generate multiple sequences by latent space manipulation on disentangled `Motion' and `Character' magnitudes ($A_c$). }
\vspace{-0.35cm}
\label{fig:a}
\end{figure}

\begin{figure}[t]
\begin{center}
\includegraphics[width=1\linewidth]{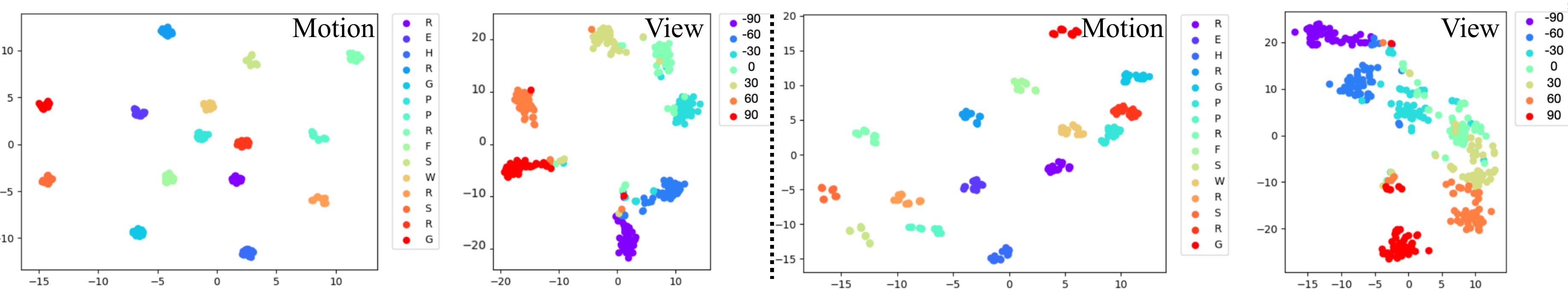}
\end{center}
\vspace{-0.35cm}
   \caption{Skeleton representations (marked by different colors with `Motion' and `View') on Mixamo with ViA by supervised (left) and unsupervised (right) motion retargeting.}
\vspace{0.cm}
\label{fig:cluster}
\end{figure}

\paragraph{Qualitative Evaluation.} To demonstrate that the `View/Subject' and `Motion' are well disentangled by the proposed framework, we visualize an example of motion retargeting inference of two Penn Action's videos (see Fig.~\ref{fig:gen}). Then we visualize the representations of all the skeletons on Mixamo with t-SNE (see Fig.~\ref{fig:cluster} with both supervised and unsupervised motion retargeting). Qualitative results validate that the `View/Subject' and `Motion' parts of 2D skeleton sequences have been effectively disentangled. To further understand the learned `Motion' features, we generate skeletons for each single input sequence with only $\mathbf{r}_{m}$ (see Fig.~\ref{fig:a} (b)), and then with $\mathbf{r}_{m}$ combined by different $\mathbf{r}_{c}$ obtained by a linearly grown $A_c$ (see Fig.~\ref{fig:a} (c)). We observe that $\mathbf{r}_{m}$ represents the motion in a `canonical view', regardless of original views of the input skeleton sequences. As such a `canonical view' can be considered as a normalized form of the given skeleton sequence, learning transformations between generative sequence and source sequence using $\mathbf{D}_c$ and $A_c$ is considerably more efficient than directly generation once the `canonical view' is fixed.

\subsection{Ablation Study}\label{sec:ablation}

To understand the contribution of each loss function in ViA, we conduct ablation experiments on Smarthome CV2 with fine-tuning protocol. To perform more studies on the characteristics of view-invariant representations, we additionally set a Cross-view (CV) action recognition protocol on the Mixamo dataset (\ie, Top-1 classification accuracy) using two different 2D skeleton projections generated by random 3D rotations of the same action for 2D cross-view evaluation.
We start from a baseline model that has been previously pre-trained on the synthetic dataset (\ie, Mixamo) using motion retargeting annotations for cross-character reconstruction. Already this visual encoder has a strong capability to embed the 2D skeleton sequence into a view-invariant representation. The results in Tab.~\ref{tab_abloss} (see L0) suggest that the generalizability is hindered by the lack of action diversity in the synthetic training dataset if directly transferring the baseline visual encoder for action classification.
Therefore, from the full results in Tab.~\ref{tab_abloss}, we infer that additional self-supervised training on Posetics can improve the real-world generalizability. Specifically, a \textbf{self reconstruction loss (L1)} can help the visual encoder learn the global characteristics of the real-world data and thus facilitate the classification. The \textbf{cycle reconstruction loss (L2)} and the \textbf{triplet loss (L3)} aim at maximizing the embedding similarity between the same action performed from two viewpoints, while minimizing the embedding similarity between different actions. These losses are instrumental in extracting a more generic representation for the downstream action classification task. Finally, The \textbf{velocity loss (L4)} contributes to minor boosts. 

\begin{table}[t]
\centering
\begin{center}
\scalebox{0.9}{

\setlength{\tabcolsep}{.6mm}{
\begin{tabular}{ l c c c c c c}
\hline
\multirow{2}*{\textbf{Methods}}  & \multirow{2}*{$\mathcal{L}_{self}$}
& \multirow{2}*{$\mathcal{L}_{cycle}$} 
& \multirow{2}*{$\mathcal{L}_{trip}$} 
& \multirow{2}*{$\mathcal{L}_{vel}$}
&{\textbf{Smarthome}}
&{\textbf{Mixamo}}
\\

&&&&& \text{CV2(\%)} &\text{CV(\%)} \\
\hline
\hline
\text{L0: Baseline}&& & &   &\text{61.7}& \text{71.7} \\
\text{L1: +Self}&\checkmark& & &  &\text{62.9}& \text{76.5} \\
\text{L2: +Cycle}&\checkmark&\checkmark& & &\text{63.8}& \text{82.5}  \\
\text{L3: +Trip}&\checkmark&\checkmark&\checkmark&  &\text{65.0}& \text{85.8}\\
\hline
\textbf{L4: +Vel (Full)}&\checkmark&\checkmark&\checkmark & \checkmark &\textbf{65.4}& \textbf{87.2}\\
\hline

\end{tabular}}}
\end{center}
\vspace{-0.4cm}
\caption{Ablation study of ViA on Smarthome CV2 and Mixamo CV with transfer learning (fine-tuning).}
\vspace{-0.35cm}
\label{tab_abloss}
\end{table}

\section{Conclusion}
In this work, we presented ViA, a generic framework aimed at learning view-invariant skeleton action representation via Latent Motion Disentanglement. We showed that self-supervised motion retargeting with contrastive learning can be an effective pretext task to learn view-invariant action representation for real-world 2D skeleton sequences. Experimental analysis confirmed that a visual encoder extracting such representation on large-scale datasets such as Posetics significantly boosts the performance when transferred onto downstream target datasets for cross-subject and cross-view action classification tasks.
Future work will extend our method to RGB data in order to capture the information on objects used to perform actions, while maintaining view- and subject-invariance. 

\bibliography{aaai23}

\begin{thebibliography}{64}
\providecommand{\natexlab}[1]{#1}

\bibitem[{Aberman et~al.(2019)Aberman, Wu, Lischinski, Chen, and
  Cohen-Or}]{2dmr}
Aberman, K.; Wu, R.; Lischinski, D.; Chen, B.; and Cohen-Or, D. 2019.
\newblock Learning character-agnostic motion for motion retargeting in 2D.
\newblock \emph{ACM TOG}.

\bibitem[{Arnab et~al.(2021)Arnab, Dehghani, Heigold, Sun, Lučić, and
  Schmid}]{arnab2021vivit}
Arnab, A.; Dehghani, M.; Heigold, G.; Sun, C.; Lučić, M.; and Schmid, C.
  2021.
\newblock ViViT: A Video Vision Transformer.
\newblock \emph{ICCV}.

\bibitem[{Bachman, Hjelm, and Buchwalter(2019)}]{bachman2019learning}
Bachman, P.; Hjelm, R.~D.; and Buchwalter, W. 2019.
\newblock Learning Representations by Maximizing Mutual Information Across
  Views.
\newblock In \emph{NeurIPS}.

\bibitem[{Caetano, Br{\'e}mond, and Schwartz(2019)}]{Caetano2019SkeletonIR}
Caetano, C.; Br{\'e}mond, F.; and Schwartz, W. 2019.
\newblock Skeleton Image Representation for {3D} Action Recognition Based on
  Tree Structure and Reference Joints.
\newblock \emph{SIBGRAPI}.

\bibitem[{Carreira and Zisserman(2017)}]{Carreira_2017_CVPR}
Carreira, J.; and Zisserman, A. 2017.
\newblock Quo Vadis, Action Recognition? A New Model and the Kinetics Dataset.
\newblock In \emph{CVPR}.

\bibitem[{Chan et~al.(2019)Chan, Ginosar, Zhou, and Efros}]{dance}
Chan, C.; Ginosar, S.; Zhou, T.; and Efros, A.~A. 2019.
\newblock Everybody Dance Now.
\newblock In \emph{ICCV}.

\bibitem[{Chen et~al.(2020)Chen, Kornblith, Norouzi, and
  Hinton}]{chen2020simclr}
Chen, T.; Kornblith, S.; Norouzi, M.; and Hinton, G. 2020.
\newblock A Simple Framework for Contrastive Learning of Visual
  Representations.
\newblock In \emph{ICML}.

\bibitem[{Chen et~al.(2021)Chen, Zhang, Yuan, Li, Deng, and
  Hu}]{topology_2021_ICCV}
Chen, Y.; Zhang, Z.; Yuan, C.; Li, B.; Deng, Y.; and Hu, W. 2021.
\newblock Channel-Wise Topology Refinement Graph Convolution for Skeleton-Based
  Action Recognition.
\newblock In \emph{ICCV}.

\bibitem[{Cheng et~al.(2020)Cheng, Zhang, He, Chen, Cheng, and
  Lu}]{shift_2020_CVPR}
Cheng, K.; Zhang, Y.; He, X.; Chen, W.; Cheng, J.; and Lu, H. 2020.
\newblock Skeleton-Based Action Recognition With Shift Graph Convolutional
  Network.
\newblock In \emph{CVPR}.

\bibitem[{Das et~al.(2019)Das, Dai, Koperski, Minciullo, Garattoni, Bremond,
  and Francesca}]{Das_2019_ICCV}
Das, S.; Dai, R.; Koperski, M.; Minciullo, L.; Garattoni, L.; Bremond, F.; and
  Francesca, G. 2019.
\newblock Toyota Smarthome: Real-World Activities of Daily Living.
\newblock In \emph{ICCV}.

\bibitem[{Duan et~al.(2022)Duan, Zhao, Chen, Shao, Lin, and
  Dai}]{duan2021revisiting}
Duan, H.; Zhao, Y.; Chen, K.; Shao, D.; Lin, D.; and Dai, B. 2022.
\newblock Revisiting Skeleton-based Action Recognition.
\newblock In \emph{CVPR}.

\bibitem[{Fang et~al.(2017)Fang, Xie, Tai, and Lu}]{fang2017rmpe}
Fang, H.-S.; Xie, S.; Tai, Y.-W.; and Lu, C. 2017.
\newblock {RMPE}: Regional Multi-person Pose Estimation.
\newblock In \emph{ICCV}.

\bibitem[{Feichtenhofer(2020)}]{x3d}
Feichtenhofer, C. 2020.
\newblock X3D: Expanding Architectures for Efficient Video Recognition.
\newblock In \emph{CVPR}.

\bibitem[{Feichtenhofer et~al.(2019)Feichtenhofer, Fan, Malik, and
  He}]{slow-fast}
Feichtenhofer, C.; Fan, H.; Malik, J.; and He, K. 2019.
\newblock SlowFast Networks for Video Recognition.
\newblock In \emph{ICCV}.

\bibitem[{Feichtenhofer, Pinz, and Zisserman(2016)}]{c-2stream}
Feichtenhofer, C.; Pinz, A.; and Zisserman, A. 2016.
\newblock Convolutional Two-Stream Network Fusion for Video Action Recognition.
\newblock In \emph{CVPR}.

\bibitem[{{Hara}, {Kataoka}, and {Satoh}(2017)}]{3d-resnet}
{Hara}, K.; {Kataoka}, H.; and {Satoh}, Y. 2017.
\newblock Learning Spatio-Temporal Features with {3D} Residual Networks for
  Actio Recognition.
\newblock In \emph{ICCVW}.

\bibitem[{He et~al.(2020)He, Fan, Wu, Xie, and Girshick}]{He_2020_CVPR}
He, K.; Fan, H.; Wu, Y.; Xie, S.; and Girshick, R. 2020.
\newblock Momentum Contrast for Unsupervised Visual Representation Learning.
\newblock In \emph{CVPR}.

\bibitem[{{He} et~al.(2015){He}, {Zhang}, {Ren}, and {Sun}}]{kaiming}
{He}, K.; {Zhang}, X.; {Ren}, S.; and {Sun}, J. 2015.
\newblock Delving Deep into Rectifiers: Surpassing Human-Level Performance on
  ImageNet Classification.
\newblock In \emph{ICCV}.

\bibitem[{Hjelm et~al.(2019)Hjelm, Fedorov, Lavoie-Marchildon, Grewal, Bachman,
  Trischler, and Bengio}]{hjelm2019learning}
Hjelm, R.~D.; Fedorov, A.; Lavoie-Marchildon, S.; Grewal, K.; Bachman, P.;
  Trischler, A.; and Bengio, Y. 2019.
\newblock Learning deep representations by mutual information estimation and
  maximization.
\newblock In \emph{ICLR}.

\bibitem[{Inc.(2018)}]{mixamo}
Inc., A.~S. 2018.
\newblock Mixamo. https://www.mixamo.com. https://www.mixamo.com.
\newblock \emph{Accessed: 2018-12-27.}

\bibitem[{Ji et~al.(2013)Ji, Xu, Yang, and Yu}]{3dcnn}
Ji, S.; Xu, W.; Yang, M.; and Yu, K. 2013.
\newblock 3D Convolutional Neural Networks for Human Action Recognition.
\newblock \emph{IEEE TPAMI}.

\bibitem[{Jiao et~al.(2020)Jiao, Xiong, Zhang, Zhang, Zhang, and
  Zhu}]{jiao2020subgraph}
Jiao, Y.; Xiong, Y.; Zhang, J.; Zhang, Y.; Zhang, T.; and Zhu, Y. 2020.
\newblock Sub-graph Contrast for Scalable Self-Supervised Graph Representation
  Learning.
\newblock In \emph{ICDM}.

\bibitem[{Karen and Andrew(2014)}]{two-stream}
Karen, S.; and Andrew, Z. 2014.
\newblock Two-stream convolutional networks for action recognition in videos.
\newblock In \emph{NeurIPS}.

\bibitem[{{Kim} and {Reiter}(2017)}]{tcn}
{Kim}, T.~S.; and {Reiter}, A. 2017.
\newblock Interpretable {3D} Human Action Analysis with Temporal Convolutional
  Networks.
\newblock In \emph{CVPRW}.

\bibitem[{Kundu et~al.(2019)Kundu, Gor, Uppala, and
  Radhakrishnan}]{encoderwacv}
Kundu, J.~N.; Gor, M.; Uppala, P.~K.; and Radhakrishnan, V.~B. 2019.
\newblock Unsupervised Feature Learning of Human Actions As Trajectories in
  Pose Embedding Manifold.
\newblock In \emph{WACV}.

\bibitem[{Li et~al.(2018)Li, Wong, Zhao, and Kankanhalli}]{view-nips2018}
Li, J.; Wong, Y.; Zhao, Q.; and Kankanhalli, M.~S. 2018.
\newblock Unsupervised Learning of View-invariant Action Representations.
\newblock In \emph{NeurIPS}.

\bibitem[{Li et~al.(2021{\natexlab{a}})Li, Li, Wang, Wang, and
  Qiao}]{li2021ctnet}
Li, K.; Li, X.; Wang, Y.; Wang, J.; and Qiao, Y. 2021{\natexlab{a}}.
\newblock CT-Net: Channel Tensorization Network for Video Classification.
\newblock In \emph{ICLR}.

\bibitem[{Li et~al.(2021{\natexlab{b}})Li, Wang, Ni, Wang, Yang, and
  Zhang}]{ConNTU}
Li, L.; Wang, M.; Ni, B.; Wang, H.; Yang, J.; and Zhang, W. 2021{\natexlab{b}}.
\newblock 3D Human Action Representation Learning via Cross-View Consistency
  Pursuit.
\newblock In \emph{CVPR}.

\bibitem[{Li et~al.(2019)Li, Chen, Chen, Zhang, Wang, and Tian}]{Li_2019_CVPR}
Li, M.; Chen, S.; Chen, X.; Zhang, Y.; Wang, Y.; and Tian, Q. 2019.
\newblock Actional-Structural Graph Convolutional Networks for Skeleton-Based
  Action Recognition.
\newblock In \emph{CVPR}.

\bibitem[{Li et~al.(2021{\natexlab{c}})Li, Liu, Zhang, Ni, Wang, and Li}]{uav}
Li, T.; Liu, J.; Zhang, W.; Ni, Y.; Wang, W.; and Li, Z. 2021{\natexlab{c}}.
\newblock UAV-Human: A Large Benchmark for Human Behavior Understanding With
  Unmanned Aerial Vehicles.
\newblock In \emph{CVPR}.

\bibitem[{{Liu} et~al.(2020){Liu}, {Shahroudy}, {Perez}, {Wang}, {Duan}, and
  {Kot}}]{NTU-120}
{Liu}, J.; {Shahroudy}, A.; {Perez}, M.; {Wang}, G.; {Duan}, L.~Y.; and {Kot},
  A.~C. 2020.
\newblock NTU RGB+D 120: A Large-Scale Benchmark for {3D} Human Activity
  Understanding.
\newblock \emph{IEEE TPAMI}.

\bibitem[{Liu et~al.(2020)Liu, Zhang, Chen, Wang, and
  Ouyang}]{liu2020disentangling}
Liu, Z.; Zhang, H.; Chen, Z.; Wang, Z.; and Ouyang, W. 2020.
\newblock Disentangling and Unifying Graph Convolutions for Skeleton-Based
  Action Recognition.
\newblock In \emph{CVPR}.

\bibitem[{Ma et~al.(2022)Ma, Rahmani, Fan, Yang, Chen, and
  Liu}]{aaai2022remote}
Ma, X.; Rahmani, H.; Fan, Z.; Yang, B.; Chen, J.; and Liu, J. 2022.
\newblock Remote: Reinforced motion transformation network for semi-supervised
  2d pose estimation in videos.
\newblock In \emph{Proceedings of the AAAI Conference on Artificial
  Intelligence}.

\bibitem[{Nie and Liu(2021)}]{nie2021view}
Nie, Q.; and Liu, Y. 2021.
\newblock View transfer on human skeleton pose: Automatically disentangle the
  view-variant and view-invariant information for pose representation learning.
\newblock \emph{IJCV}.

\bibitem[{Nie, Liu, and Liu(2020)}]{nie2020unsupervised}
Nie, Q.; Liu, Z.; and Liu, Y. 2020.
\newblock Unsupervised Human {3D} Pose Representation with Viewpoint and Pose
  Disentanglement.
\newblock In \emph{ECCV}.

\bibitem[{Rogez, Weinzaepfel, and Schmid(2019)}]{RogezWS18}
Rogez, G.; Weinzaepfel, P.; and Schmid, C. 2019.
\newblock {LCR-Net++: Multi-person {2D} and {3D} Pose Detection in Natural
  Images}.
\newblock \emph{IEEE TPAMI}.

\bibitem[{Ryoo et~al.(2020)Ryoo, Piergiovanni, Kangaspunta, and
  Angelova}]{Ryoo2020AssembleNetAM}
Ryoo, M.; Piergiovanni, A.; Kangaspunta, J.; and Angelova, A. 2020.
\newblock AssembleNet++: Assembling Modality Representations via Attention
  Connections.
\newblock \emph{ECCV}.

\bibitem[{Sardari, Ommer, and Mirmehdi(2021)}]{bmvc2021unsupervised}
Sardari, F.; Ommer, B.; and Mirmehdi, M. 2021.
\newblock Unsupervised View-Invariant Human Posture Representation.
\newblock In \emph{BMVC}.

\bibitem[{Shahroudy et~al.(2016)Shahroudy, Liu, Ng, and
  Wang}]{Shahroudy2016NTURA}
Shahroudy, A.; Liu, J.; Ng, T.-T.; and Wang, G. 2016.
\newblock NTU RGB+D: A Large Scale Dataset for {3D} Human Activity Analysis.
\newblock \emph{CVPR}.

\bibitem[{Shi et~al.(2019)Shi, Zhang, Cheng, and Lu}]{2sagcn2019cvpr}
Shi, L.; Zhang, Y.; Cheng, J.; and Lu, H. 2019.
\newblock Two-Stream Adaptive Graph Convolutional Networks for Skeleton-Based
  Action Recognition.
\newblock In \emph{CVPR}.

\bibitem[{Skorokhodov, Tulyakov, and Elhoseiny(2022)}]{stylegan_v}
Skorokhodov, I.; Tulyakov, S.; and Elhoseiny, M. 2022.
\newblock StyleGAN-V: A Continuous Video Generator with the Price, Image
  Quality and Perks of StyleGAN2.
\newblock In \emph{CVPR}.

\bibitem[{Song et~al.(2020)Song, Zhang, Shan, and Wang}]{res-gcn}
Song, Y.-F.; Zhang, Z.; Shan, C.; and Wang, L. 2020.
\newblock Stronger, Faster and More Explainable: A Graph Convolutional Baseline
  for Skeleton-Based Action Recognition.
\newblock In \emph{ACM MM}.

\bibitem[{Su, Lin, and Wu(2021)}]{motionconsistency}
Su, Y.; Lin, G.; and Wu, Q. 2021.
\newblock Self-Supervised 3D Skeleton Action Representation Learning With
  Motion Consistency and Continuity.
\newblock In \emph{ICCV}.

\bibitem[{Sun et~al.(2020)Sun, Zhao, Chen, Schroff, Adam, and
  Liu}]{sun2020viewinvariant}
Sun, J.~J.; Zhao, J.; Chen, L.-C.; Schroff, F.; Adam, H.; and Liu, T. 2020.
\newblock View-Invariant Probabilistic Embedding for Human Pose.
\newblock In \emph{ECCV}.

\bibitem[{Tian, Krishnan, and Isola(2020)}]{tian2020cmc}
Tian, Y.; Krishnan, D.; and Isola, P. 2020.
\newblock Contrastive Multiview Coding.
\newblock In \emph{ECCV}.

\bibitem[{Tulyakov et~al.(2018)Tulyakov, Liu, Yang, and
  Kautz}]{Tulyakov:2018:MoCoGAN}
Tulyakov, S.; Liu, M.-Y.; Yang, X.; and Kautz, J. 2018.
\newblock {MoCoGAN}: Decomposing motion and content for video generation.
\newblock In \emph{CVPR}.

\bibitem[{Vemulapalli, Arrate, and Chellappa(2014)}]{Vemulapalli2014HumanAR}
Vemulapalli, R.; Arrate, F.; and Chellappa, R. 2014.
\newblock Human Action Recognition by Representing {3D} Skeletons as Points in
  a Lie Group.
\newblock \emph{CVPR}.

\bibitem[{Villegas et~al.(2018)Villegas, Yang, Ceylan, and Lee}]{NKN_2018_CVPR}
Villegas, R.; Yang, J.; Ceylan, D.; and Lee, H. 2018.
\newblock Neural Kinematic Networks for Unsupervised Motion Retargetting.
\newblock In \emph{CVPR}.

\bibitem[{Wang et~al.(2014)Wang, Nie, Xia, Wu, and Zhu}]{ucla_2014_CVPR}
Wang, J.; Nie, X.; Xia, Y.; Wu, Y.; and Zhu, S.-C. 2014.
\newblock Cross-view Action Modeling, Learning and Recognition.
\newblock In \emph{CVPR}.

\bibitem[{Wang et~al.(2021)Wang, Tong, Ji, and Wu}]{Wang_2021_CVPR}
Wang, L.; Tong, Z.; Ji, B.; and Wu, G. 2021.
\newblock TDN: Temporal Difference Networks for Efficient Action Recognition.
\newblock In \emph{CVPR}.

\bibitem[{Wang et~al.(2020)Wang, Bilinski, Bremond, and
  Dantcheva}]{Wang_2020_CVPR}
Wang, Y.; Bilinski, P.; Bremond, F.; and Dantcheva, A. 2020.
\newblock {G3AN}: Disentangling Appearance and Motion for Video Generation.
\newblock In \emph{CVPR}.

\bibitem[{WANG et~al.(2020)WANG, Bilinski, Bremond, and
  Dantcheva}]{WANG_2020_WACV}
WANG, Y.; Bilinski, P.; Bremond, F.; and Dantcheva, A. 2020.
\newblock Ima{GIN}ator: Conditional Spatio-Temporal GAN for Video Generation.
\newblock In \emph{WACV}.

\bibitem[{Wang et~al.(2022)Wang, Yang, Bremond, and Dantcheva}]{wang2022latent}
Wang, Y.; Yang, D.; Bremond, F.; and Dantcheva, A. 2022.
\newblock Latent Image Animator: Learning to Animate Images via Latent Space
  Navigation.
\newblock In \emph{ICLR}.

\bibitem[{Weinzaepfel and Rogez(2021)}]{weinzaepfel2021mimetics}
Weinzaepfel, P.; and Rogez, G. 2021.
\newblock Mimetics: Towards Understanding Human Actions Out of Context.
\newblock \emph{IJCV.}

\bibitem[{{Wu} et~al.(2018){Wu}, {Xiong}, {Yu}, and {Lin}}]{non-para}
{Wu}, Z.; {Xiong}, Y.; {Yu}, S.~X.; and {Lin}, D. 2018.
\newblock Unsupervised Feature Learning via Non-parametric Instance
  Discrimination.
\newblock In \emph{CVPR}.

\bibitem[{Yan, Xiong, and Lin(2018)}]{Yan2018SpatialTG}
Yan, S.; Xiong, Y.; and Lin, D. 2018.
\newblock Spatial Temporal Graph Convolutional Networks for Skeleton-Based
  Action Recognition.
\newblock \emph{AAAI}.

\bibitem[{Yang et~al.(2021{\natexlab{a}})Yang, Dai, Wang, Mallick, Minciullo,
  Francesca, and Bremond}]{Yang_2021_WACV}
Yang, D.; Dai, R.; Wang, Y.; Mallick, R.; Minciullo, L.; Francesca, G.; and
  Bremond, F. 2021{\natexlab{a}}.
\newblock Selective Spatio-Temporal Aggregation Based Pose Refinement System:
  Towards Understanding Human Activities in Real-World Videos.
\newblock In \emph{WACV}.

\bibitem[{Yang et~al.(2021{\natexlab{b}})Yang, Wang, Dantcheva, Garattoni,
  Francesca, and Bremond}]{orvpe}
Yang, D.; Wang, Y.; Dantcheva, A.; Garattoni, L.; Francesca, G.; and Bremond,
  F. 2021{\natexlab{b}}.
\newblock Self-Supervised Video Pose Representation Learning for
  Occlusion-Robust Action Recognition.
\newblock In \emph{FG}.

\bibitem[{Yang et~al.(2021{\natexlab{c}})Yang, Wang, Dantcheva, Garattoni,
  Francesca, and Bremond}]{unik}
Yang, D.; Wang, Y.; Dantcheva, A.; Garattoni, L.; Francesca, G.; and Bremond,
  F. 2021{\natexlab{c}}.
\newblock UNIK: A Unified Framework for Real-world Skeleton-based Action
  Recognition.
\newblock In \emph{BMVC}.

\bibitem[{Yang et~al.(2021{\natexlab{d}})Yang, Liu, Lu, Er, and
  Kot}]{colorization}
Yang, S.; Liu, J.; Lu, S.; Er, M.~H.; and Kot, A.~C. 2021{\natexlab{d}}.
\newblock Skeleton Cloud Colorization for Unsupervised 3D Action Representation
  Learning.
\newblock In \emph{ICCV}.

\bibitem[{Yu et~al.(2022)Yu, Tack, Mo, Kim, Kim, Ha, and Shin}]{yu2022digan}
Yu, S.; Tack, J.; Mo, S.; Kim, H.; Kim, J.; Ha, J.-W.; and Shin, J. 2022.
\newblock Generating Videos with Dynamics-aware Implicit Generative Adversarial
  Networks.
\newblock In \emph{ICLR}.

\bibitem[{{Zhang}, {Zhu}, and {Derpanis}(2013)}]{penn}
{Zhang}, W.; {Zhu}, M.; and {Derpanis}, K.~G. 2013.
\newblock From Actemes to Action: A Strongly-Supervised Representation for
  Detailed Action Understanding.
\newblock In \emph{ICCV}.

\bibitem[{Zhao et~al.(2021)Zhao, Wang, Zhao, Yuan, Sun, Schroff, Adam, Peng,
  Metaxas, and Liu}]{cv-mim}
Zhao, L.; Wang, Y.; Zhao, J.; Yuan, L.; Sun, J.~J.; Schroff, F.; Adam, H.;
  Peng, X.; Metaxas, D.; and Liu, T. 2021.
\newblock Learning View-Disentangled Human Pose Representation by Contrastive
  Cross-View Mutual Information Maximization.
\newblock In \emph{CVPR}.

\bibitem[{Zhu et~al.(2017)Zhu, Park, Isola, and Efros}]{CycleGAN2017}
Zhu, J.-Y.; Park, T.; Isola, P.; and Efros, A.~A. 2017.
\newblock Unpaired Image-to-Image Translation using Cycle-Consistent
  Adversarial Networks.
\newblock In \emph{ICCV}.

\end{thebibliography}
\end{document}